\documentclass[journal]{IEEEtran}
\usepackage{graphicx,float,subfig,epstopdf}
\usepackage{xcolor}
\usepackage{multirow}
\usepackage{tabularx}
\usepackage{url}
\usepackage{booktabs}
\usepackage{lineno}
\usepackage[noadjust]{cite}
\usepackage{hyperref}
\usepackage{changes}
\ifCLASSINFOpdf
\else
\fi
\begin{document}

%\linenumbers
%\linenumbersep 3pt\relax

%\title{Anatomical Priors for Image Segmentation via Post-Processing with Denoising Autoencoders}
\title{Post-DAE: Anatomically Plausible Segmentation \\via Post-Processing with Denoising Autoencoders}

%\author{
%\IEEEauthorblockN{Agostina J Larrazabal$^1$,
%C\'esar Mart\'inez$^1$, Ben Glocker$^2$ and Enzo Ferrante$^1$}

%\IEEEauthorblockA{$^1$\textit{Research
%Institute for Signals, Systems and Computational Intelligence, sinc(i) (FICH-UNL, CONICET)}\\Santa Fe, Argentina\\}
%\IEEEauthorblockA{{$^2$\textit{Biomedical Image Analysis Group, Imperial College London}\\
%London, UK\\}
%}

\author{Agostina J Larrazabal,
        C\'esar Mart\'inez,
         Ben Glocker
        and Enzo Ferrante% <-this % stops a space
\thanks{Article accepted for publication in IEEE Transactions on Medical Imaging. Copyright (c) 2020 IEEE. Personal use of this material is permitted. However, permission to use this material for any other purposes must be obtained from the IEEE by sending a request to pubs-permissions@ieee.org.

A.J. Larrazabal, C. Mart\'inez and E. Ferrante are with the Institute for Signals, Systems and Computational Intelligence, sinc(i) CONICET-UNL, Santa Fe, Argentina. (e-mails: alarrazabal@sinc.unl.edu.ar - cmartinez@sinc.unl.edu.ar, eferrante@sinc.unl.edu.ar). B. Glocker is with the Biomedical Image Analysis Group, Imperial College London, London, UK. (e-mail: b.glocker@imperial.ac.uk)}

\thanks{E. Ferrante is beneficiary of an AXA Research Fund grant. The authors gratefully acknowledge NVIDIA Corporation with the donation of the GPUs used for this research, and the support of UNL (CAID-PIC-50420150100098LI, CAID-PIC-50220140100084LI) and ANPCyT (PICT 2016-0651, PICT 2018-03907).}
}

\maketitle

\begin{abstract}
We introduce Post-DAE, a post-processing method based on denoising autoencoders (DAE) to improve the anatomical plausibility of arbitrary biomedical image segmentation algorithms. Some of the most popular segmentation methods (e.g. based on convolutional neural networks or random forest classifiers) \textcolor{black}{incorporate additional post-processing steps to ensure that the resulting masks fulfill expected connectivity constraints}. \textcolor{black}{These methods operate under the hypothesis that contiguous pixels with similar aspect should belong to the same class}. \textcolor{black}{Even if valid in general, this assumption does not consider more complex priors like topological restrictions or convexity, which cannot be easily incorporated into these methods.} %Moreover, they usually increase the overall time complexity of the complete segmentation pipeline.

Post-DAE leverages the latest developments in manifold learning via denoising autoencoders. \textcolor{black}{First, we learn a compact and non-linear embedding that represents the space of anatomically plausible segmentations. Then, given a segmentation mask obtained with an arbitrary method, we reconstruct its anatomically plausible version by projecting it onto the learnt manifold.} \textcolor{black}{The proposed method is trained using unpaired segmentation mask, what makes it independent of intensity information and image modality.} %This enables the use of anatomical segmentations that do not need to be paired with intensity images from the same modality.} 
We performed experiments in binary and multi-label segmentation of chest X-ray \textcolor{black}{and cardiac magnetic resonance} images. \textcolor{black}{We show how erroneous and noisy segmentation masks can be improved using Post-DAE. With almost no additional computation cost, our method brings erroneous segmentations back to a feasible space.}

\end{abstract}


\begin{IEEEkeywords}
anatomical segmentation, autoencoders, convolutional neural networks, learning representations, post-processing
\end{IEEEkeywords}

% \fi
%\IEEEpeerreviewmaketitle

\section{Introduction}

\IEEEPARstart{A}{natomical} segmentation is a fundamental task in medical image computing, which consists in associating pixels of a medical image with a given organ or anatomical structure. \textcolor{black}{}It constitutes an essential step in many imaging pipelines such as computer assisted diagnosis, morphometric analysis for population studies and radiotherapy planning. \textcolor{black}{The correctness and anatomical plausibility of these results is thus of paramount importance}, since it will directly influence the overall quality of subsequent analyses.

\textcolor{black}{Convolutional neural networks (CNNs) proved to perform biomedical image segmentation in a highly accurate way} \cite{RonnebergerUnet15,KamnitsasDeepmedic16,Shakeri2016}. CNNs constitute a particular type of neural network specially suited for regularly structured data, like 2D or 3D images, where hierarchical representations of the input are learned using stacked convolutional layers. \textcolor{black}{At every layer, shared parameters (also referred as weights or kernel) are used to learn new representations of the input image. This sharing scheme reduces the number of parameters that should be learnt and allows the use of CNNs in large images.} Thanks to the inherently regular structure of the images, these parameters are successively convoluted with the input data resulting in more abstract representations. \textcolor{black}{This trick is particularly helpful for tasks in which invariance to translation is an expected property, such as image classification. However, in medical images, where the location of the anatomical structures is often highly regular, this property may lead to incorrect predictions in regions with similar intensities when insufficient contextual information is considered}. \textcolor{black}{The organs observed in anatomical images tend to preserve shape and topology across patients. Nonetheless, the pixel-level predictions of most CNN architectures do not account for such higher-order topological properties, as discussed in \cite{bentaieb2016topology}}.

\textcolor{black}{Before the emergence of CNNs, other learning-based methods were popular for biomedical image segmentation (e.g. Random Forest (RF) \cite{breiman2001random}). When the amount of annotated data is small and insufficient for training deep CNNs, some of these classical methods are still in use. A popular strategy is to adopt patch-based methods, where handcrafted features are generated from image patches and then used to train a classifier. Such classifier will then make pixel-level predictions considering only the image area around the central pixel of the patch. This results in methods which are also agnostic to the global shape and topology of the anatomical structures.}

\begin{figure}[t!]
   \includegraphics[width=\columnwidth]{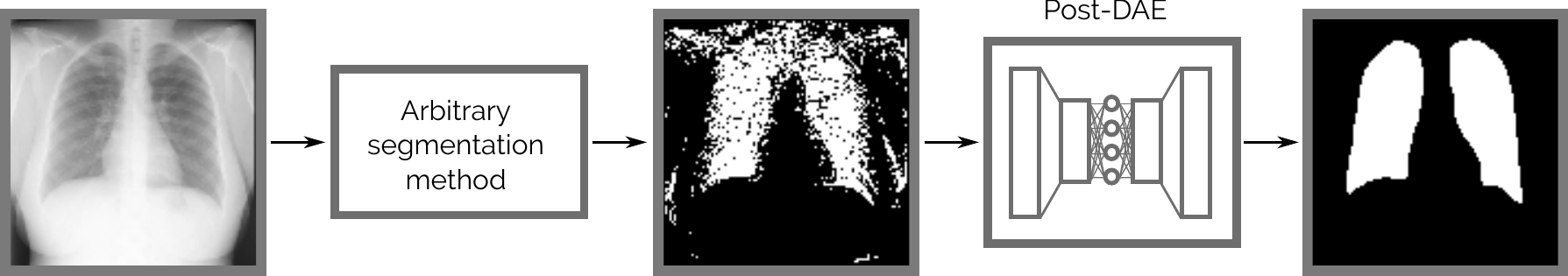}
    \caption{\textcolor{black}{Post-DAE worflow: the method is implemented as a post-processing step which maps arbitrary segmentation masks to anatomical plausible cases.}}
\label{fig:workflow}   
\end{figure}

\textcolor{black}{In this work we introduce Post-DAE, a post-processing method which improves pixel-level predictions generated with arbitrary classifiers by incorporating shape and topological priors. We employ denoising autoencoders (DAEs) to learn compact and non-linear representations of anatomical structures, using only segmentation masks. The DAE is then used to bring potentially erroneous segmentation masks into an anatomically plausible space }(see Figure \ref{fig:workflow}). \\

\noindent \textbf{Contributions.} {A preliminary version of this work was published in MICCAI 2019 \cite{Larrazabal2019}. In this extended version we provide \textcolor{black}{additional experiments in the context of multi-class lung and heart segmentation of X-ray images, and left ventricle delineation in cardiac magnetic resonance (CMR) images. We also include a more complete and updated state-of-the-art section,
% an extended validation using segmentation-only datasets, 
a deeper analysis of how our method behaves in images with gross abnormalities and out-of-distribution cases, together with additional illustrations and extended discussion}. 

Our contributions can be summarized as follows: (i) we show that denoising autoencoders \textcolor{black}{used as a post-processing step can improve the anatomical plausibility of unfeasible segmentation masks}; (ii) we present results in the context of binary and multi-label segmentation of \textcolor{black}{chest X-ray and CMR images}, bench-marking with other classical post-processing method and showing the robustness of Post-DAE by improving segmentation masks coming from both, CNN and RF-based classifiers and \textcolor{black}{(iii) we analyze the behaviour and limitations of our method when post-processing abnormal and out-of-distribution anatomical segmentation masks.}
%(iii) we show that Post-DAE can be trained with unpaired segmentation masks annotated on different image modalities or coming from segmentation-only datasets, highlighting the fact that our method does not require image intensity information for training/testing, making it robust to domain shift
%\textcolor{black}{and (iv) we evaluate the behaviour of the porposed Post-DAE when analysis abnormal segmentations.} %Our method produces substantial improvement in non-realistic segmentations, while refining the results when segmentations are already of acceptable quality.\\
%, enabling the flexibility of having an external description of the anatomy that need not be available in the current data

\section{Related Work}

Multiple alternatives have been proposed to incorporate prior knowledge in medical image segmentation (see \cite{nosrati2016incorporating} for a complete review). 
%One option is to tackle anatomical segmentation by combining it with image registration \cite{mansilla2020,shakeri2016prior,lee2019tetris}, explicitly enforcing shape priors by deforming existing segmentation masks. However, in this work, we will focus on methods which directly learn to perform image segmentation and do not really on additional registration steps. 
One popular strategy to integrate priors about shape and topology into learning based segmentation methods is to modify the loss function used to train the model. \textcolor{black}{A topology aware loss function which incorporates high-order regularization was proposed in \cite{bentaieb2016topology}. In this case, a manually defined topological validity table specifies the relation between the structures of interest. This constitutes a disadvantage since such loss function must be constructed ad-hoc for every dataset. More similar to our method are those by \cite{Oktay2017,Ravishankar2017}, where compact anatomical representions are learnt by means of autoencoders. Such global representation is incorporated into the loss function and used to encourage anatomical plausibility into the predicted segmentation masks. Differently from our method designed to improve arbitrary segmentations, the main disadvantage of  \cite{Oktay2017,Ravishankar2017} is that they are specifically tailored to be used when training a CNN model. Therefore, they cannot be used to improve results obtained with other segmentation approaches like RF or even level-sets methods, which do not rely on an explicit training phase.}

{\color{black}An alternative simple but effective approach is to increase the receptive field of the network, i.e. the area of the input image that influences a single prediction. Even if this strategy does not incorporate an explicit shape prior, it allows the network to consider high-order interactions between distant image regions, learning to encode certain global features about shape and topology. In this regard, \cite{KamnitsasDeepmedic16} proposed to increase the receptive field of the CNN by means of a dual path focusing on a wider low resolution area of the input image. This increases the contextual information provided to the network, but also augments the complexity of the segmentation model itself. Another approach to deal with the lack of spatial context in patch-based convolutional architectures is to augment the model including information about pixel location. In \cite{wachinger2014importance} the authors suggest that location information is a crucial discriminator in patch-based image segmentation, and show experimental results about the gain in performance when adding it explicitly to the description. In a more recent work \cite{wachinger2018deepnat}, the authors propose the use of a spectral location parametrization specially adapted to brain volume coordinates to improve CNN-based image segmentation. However, albeit the fact that these strategies increase the accuracy of the resulting segmentation masks by incorporating contextual information, they do not incorporate explicit priors about shape and topology.}

\textcolor{black}{Alternative approaches implemented as post-processing methods have also been considered. Shakeri and co-workers \cite{Shakeri2016} pose the problem as a discrete energy minimization problem, where CNN predictions are seen as unary potentials of a Markov random field (MRF) \cite{Paragios2016}. In this framework, pairwise relations are used to propagate spatial homogeneity. Following a similar idea, a fully connected conditional random field (CRF) is used in \cite{KamnitsasDeepmedic16} as a post-processing step. However, as stated by \cite{KamnitsasDeepmedic16}, it is hard to find a global set of CRF parameters which can consistently improve the segmentation of all structures of interest. Moreover, there is no shape or topological prior incorporated in these models. Instead, these methods only operate under the hypothesis that pixels which are contiguous and exhibit similar aspect should belong to the same class. Even if valid in general, this assumption does not consider more complex priors like topological restrictions or convexity, which can be easily encoded in our post-processing methods. }

{\color{black}Similar to the work of \cite{dalca2018anatomical}, our model can be trained using segmentation-only datasets which are not paired with image data (or data of the same image modality as the problem at hand). Our model is agnostic to image intensity (and thus insensitive to domain shift) and its formulation is much simpler than the one introduced in \cite{dalca2018anatomical}.}

\section{\textcolor{black}{Anatomically Plausible Segmentation via Post-DAE}}

\begin{figure*}[t!]
\centering
   \includegraphics[width=0.98\textwidth]{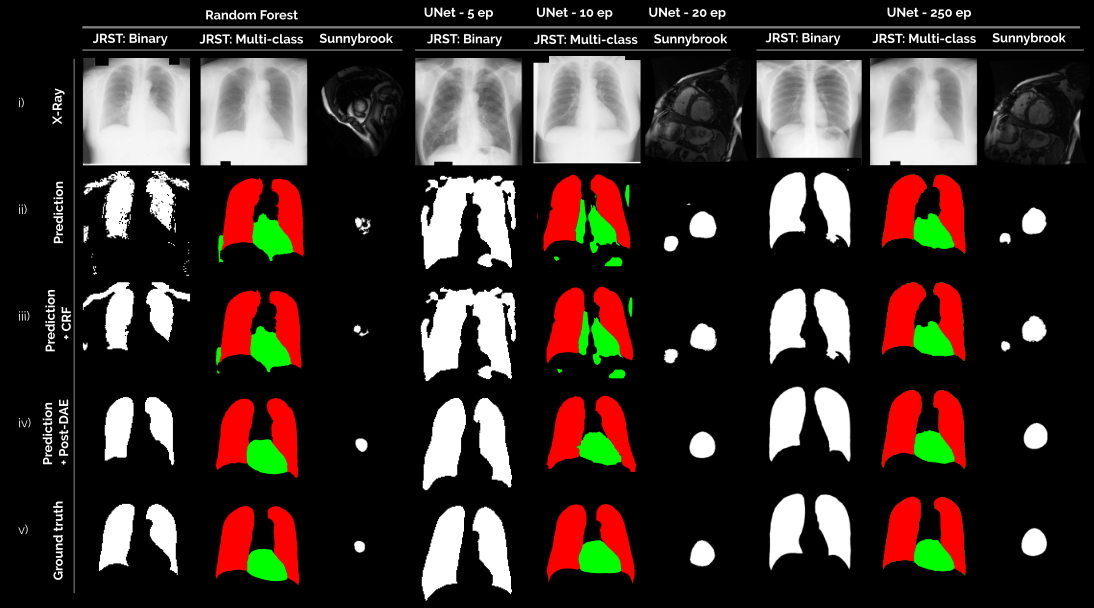}
    \caption{{\color{black}Predictions obtained with segmentation methods of several qualities: random forest and UNet trained for different number of epochs. We include examples for both binary (white images) and multi-class (color images) segmentation.} \textcolor{black}{for: (i) {\color{black}X-Ray or CMR image}; (ii) segmentation mask predicted by each baseline method; (iii) segmentation mask after post-processing with a CRF; (iv) segmentation mask after post-processing with our Post-DAE; and (v) ground-truth expert segmentations.} }
\label{fig:qualitativeResults}   
\end{figure*}
%\subsection{Problem statement.}
\textcolor{black}{Given a dataset of anatomical segmentation masks (without paired intensity images) $\mathcal{D_A} = \{S^A_i\}_{0 \leq i \leq |\mathcal{D_A}|}$ we intend to learn a model that can bring segmentations $\mathcal{D_P} = \{S^P_i\}_{0 \leq i \leq |\mathcal{D_P}|}$ predicted by different classifiers $P$ into an anatomically feasible space.} We stress the fact that our method works as a post-processing step in the space of segmentations, making it independent of the predictor, image intensities and modality.% \textcolor{black}{We apply DAEs to learn such model.}

%\subsection{Denoising autoencoders.} 
\textcolor{black}{Denoising autoenconders (DAE) are neural networks designed to reconstruct a clean input from a  corrupted version of it \cite{Vincent2010}.} \textcolor{black}{ In this work, we propose to employ DAEs to recover anatomically plausible segmentation masks from corrupted or incorrect ones.} The standard architecture for an autoencoder follows an encoder-decoder scheme (see the \textcolor{black}{Supplementary Material} for a detailed description of the architecture used in this work). \textcolor{black}{The encoder $f_{enc}(S_i)$ is a mapping function that turns the input into a lower dimensional hidden encoding $h$. In our implementation, $f_{enc}(S_i)$ is composed of stacked convolutions, non-linearities and pooling layers. At the end, a fully connected layer concentrates all information into a low dimensional code $h$. Then, the decoder $f_{dec}(h)$ maps this code back to the original input dimensions by means of successive non-linearities and up-convolutions.  }

\textcolor{black}{The model is called \textit{denosing} autoenconder due to the fact that it is trained with noisy segmentations $\hat{S}_i = \phi(S_i)$, which are obtained by degrading the ground-truth segmentation masks with a degradation function $\phi$. To minimize the reconstruction error of the predicted segmentations with respect to the ground-truth, we train the DAE using a loss function based on the Dice coefficient (DSC) (for an exhaustive description of the Dice loss please see \cite{milletari2016v} ):
\begin{equation}
    \mathcal{L}_{DAE}(S_i) = DSC(S_i, f_{dec}(f_{enc}(\phi(S_i))).
\end{equation}
The learnt encoding $h = f_{enc}(S_i)$ is forced to retain as much information as possible about the input. This is due to the bottleneck effect produced by the reduced dimensionaliy of $h$.}   \textcolor{black}{In this context, minimizing the reconstruction loss amounts to maximizing a lower bound on the mutual information between input $S_i$ and the learnt representation $h$} \cite{Vincent2010}.

\subsection{Mask degradation strategy.}
\label{sec:maskdegradation}
\textcolor{black}{We simulate corrupted segmentations to train the DAE by artificially degrading the ground truth segmentation masks $S_i$ with the following random degradation functions $\phi(S_i)$\footnote{Our code associated to Post-DAE, the degradation function and the UNet model is publicly available at the following Colab Python Notebook:\url{https://colab.research.google.com/drive/1wLZLo81clNR_c-UJTpBV4fh-BMbByAU8?usp=sharing}}: (i) We simulate over and under segmentation by adding and removing random geometric shapes (including polygons, lines and ellipses); (ii) erosion, dilation and other morphological operations with variable kernels are applied to perform minor mask alterations; (iii) mask borders are modified by random swapping of foreground-background labels in the pixels close to the organ boundaries. }} In addition, data augmentation was performed by randomly resizing the original masks.
 
\begin{figure*}[t!]
\begin{center}
   \includegraphics[width=\textwidth]{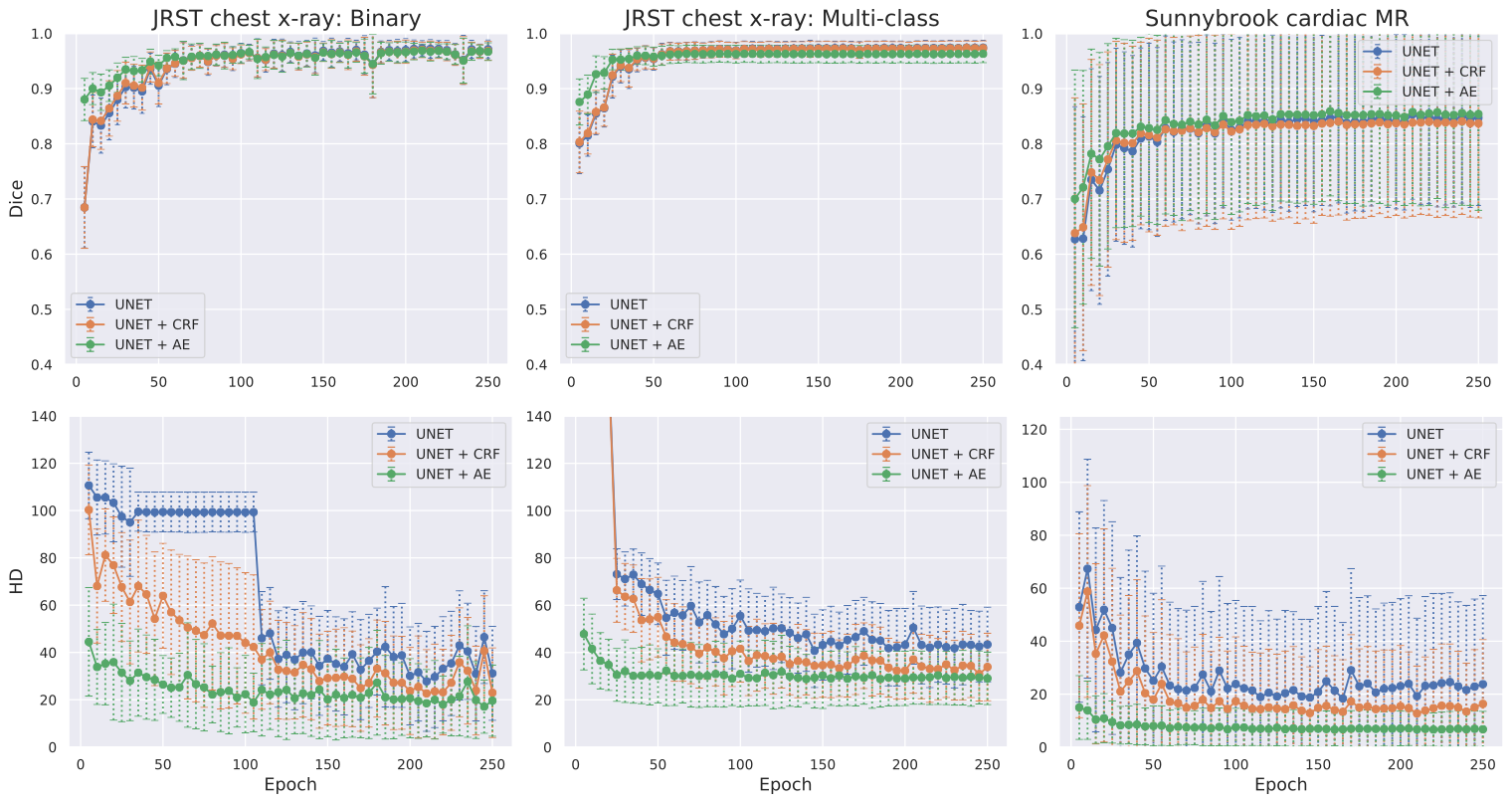}
    \caption{Quantitative comparison between post-processing with Post-DAE and CRF \cite{krahenbuhl2011efficient}. 
    %adopted as post-processing step by many segmentation methods like \cite{KamnitsasDeepmedic16}. 
    We show mean and standard deviation for post-processing UNet predictions on the test fold at different training stages (from 5 epochs to convergence). {\color{black} First column shows the results for binary lung segmentation in JRST, the second one for multi-class (lung, heart) in JSRT and the third one for binary LV masks in Sunnybrook.}
     We use Dice coefficient and Hausdorff distance (HD) to measure the segmentation quality. Note that initial low quality segmentations are improved both in terms of Dice and HD. For better initial segmentations, significant improvements in terms of HD are still obtained.  }
     %The symbol $*$ indicates that Post-DAE outperforms the other methods (no post-processing and CRF) with statistical significance (p-value $<$ 0.05 according to Wilcoxon test). The green triangle in the box indicates the mean value.
    
\label{fig:quantitativeResults}
\end{center}
\end{figure*}

\subsection{Post-processing with denoising autoencoders.} \textcolor{black}{The proposed method is rooted in the so-called manifold assumption \cite{chapelle2009semi}, which states that natural high dimensional data (like anatomical segmentation masks) concentrate close to a non-linear low-dimensional manifold.} \textcolor{black}{We use the DAE to learn such anatomically plausible manifold.} \textcolor{black}{ Then, given a segmentation mask $S^P_i$ produced with an arbitrary predictor $P$ (e.g. CNN or RF), we project it onto that manifold using $f_{enc}$ and reconstruct the corresponding anatomically feasible mask with $f_{dec}$.} \textcolor{black}{ Different from other approaches like \cite{Oktay2017,Ravishankar2017} which incorporate the anatomical priors during the segmentation network training, our method is agnostic to the training process of the original predictor. Since it is conceived as a post-processing step, segmentation masks produced with arbitrary methods can by improved using Post-DAE.}
{\color{black}Recent studies \cite{pawlowski2018unsupervised,uzunova2019unsupervised} show that different autoencoders trained with healthy brain images (operating in the intensity domain) can be used to perform anomaly detection on pathological brain images, by just looking at the differences between the original pathological image and the one processed by the autoencoder. In the same spirit, }our hypothesis (empirically validated with the experiments presented in the next section) is that those masks which are far from the anatomical space, will be mapped to a similar, but anatomically plausible segmentation. \textcolor{black}{Meanwhile, masks which are anatomically correct, will suffer almost no modification, being mapped to themselves.}

\begin{figure*}[h]
\begin{center}
     \includegraphics[width=\textwidth]{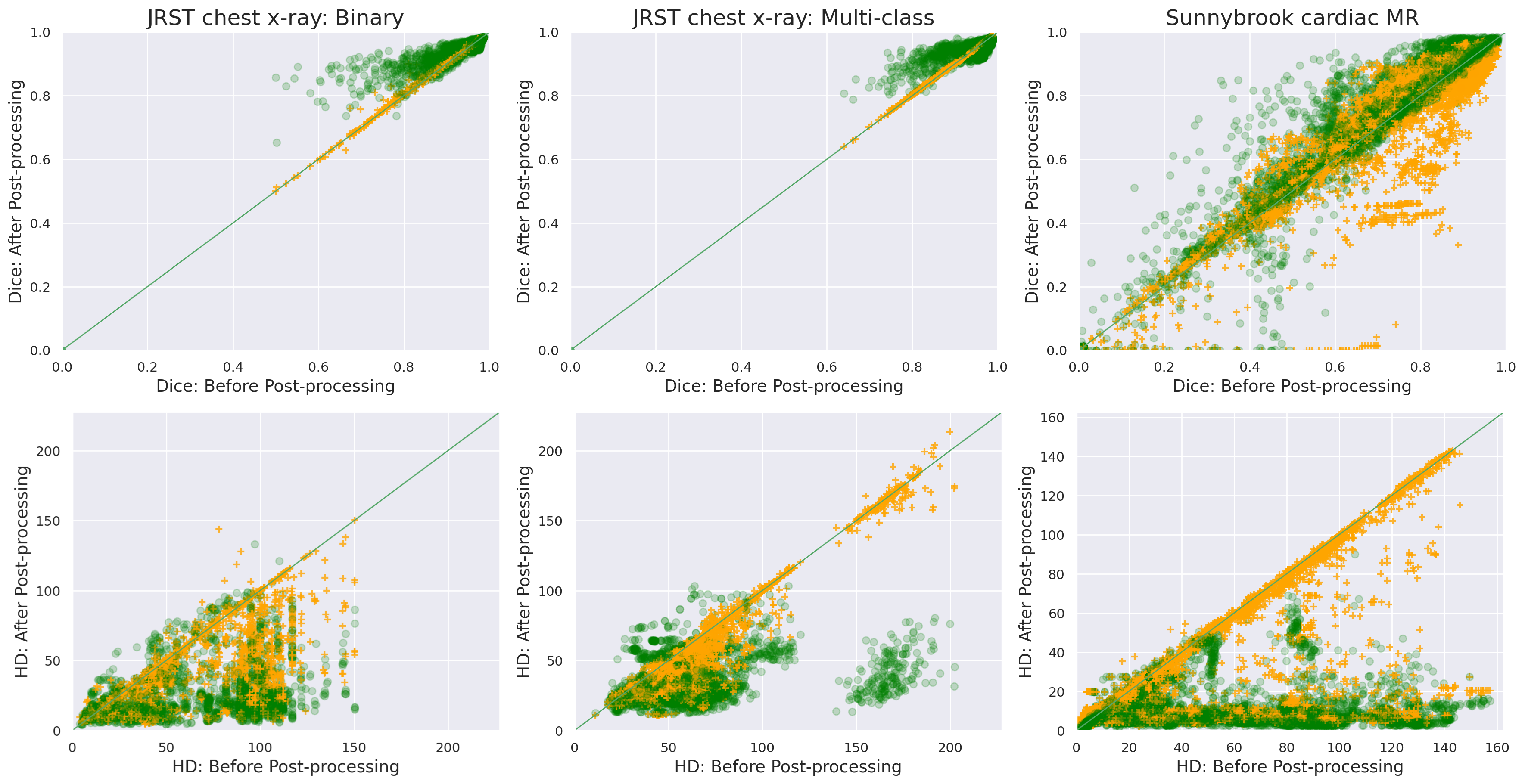}
    \caption{\textcolor{black}{Scatter plots comparing Dice (top) and HD (bottom) before and after post-processing with Post-DAE (green) and CRF (orange) for all samples in the previous study. We include segmentation masks generated with the UNet models trained from 5 to 250 epochs and random forest. First column shows the results for binary lung segmentation in JRST, the second one for multi-class (lung, heart) in JSRT and the third one for binary LV masks in Sunnybrook.}}
\label{fig:scatter}   
\end{center}
\end{figure*}
\section{Experimental setting}
\subsection{Database description.} We benchmark the proposed method \textcolor{black}{in two different anatomical segmentation scenarios, including chest X-ray and cardiac magnetic resonance (CMR) images.}

\noindent \textbf{\textcolor{black}{Chest X-ray dataset:}} in the context of lung \textcolor{black}{and heart} segmentation in X-Ray images, \textcolor{black}{we used} the Japanese Society of Radiological Technology (JSRT) database \cite{JSRT}. \textcolor{black}{This is a public dataset with expert annotations composed of 247 PA thoracic X-ray images (2048x2048 pixels and spacing of 0.175mm x 0.175mm), which are downsampled to 1024x1024 in our experiments.  Lungs \textcolor{black}{and heart} present high inter-subject variability, what makes the representation learning task especially challenging.} \textcolor{black}{We divide the database in 3 folds considering 70\% for training, 10\% for validation and 20\% for testing.} The same folds were used to train the U-Net, random forest and Post-DAE methods. \textcolor{black}{We did not apply image alignment for pre-processing.}

\noindent \textcolor{black}{\textbf{Cardiac MR dataset:} We used images from a version of the Sunnybrook Cardiac Dataset (SCD) \cite{SCD} publicly available at \url{https://github.com/mshunshin/SegNetCMR}. It includes 45 cine-MR images (every image composed of 6 to 12 short-axis (SAX) 2D slices) captured at end-systole (ES) and end-diastole (ED) time points, with corresponding segmentation masks of the left ventricle (LV). The image resolution is 256x256, covering a field of view of 320 mm x 320 mm. We partitioned the dataset using the originally suggested train/test partition scheme (taking 35 images from the training fold for validation).}

\subsection{Post-processing with CRF.}
\textcolor{black}{The proposed method is compared with a standard post-processing strategy based on a fully connected CRF \cite{krahenbuhl2011efficient}.} \textcolor{black}{This method operates under the hypothesis that pixels which are contiguous and exhibit similar aspect should belong to the same class. We use an efficient implementation of a dense CRF. }\footnote{We used the public implementation available at \url{https://github.com/lucasb-eyer/pydensecrf} with Potts compatibility function and hand-tuned parameters $\theta_\alpha=17$, $\theta_\beta=3$, $\theta_\gamma=3$ for \textcolor{black}{the X-ray images and $\theta_\alpha=7$, $\theta_\beta=3$, $\theta_\gamma=3$ for the CMR images,} chosen using the validation fold. See the website for more details about the aforementioned parameters.} \textcolor{black}{Since the CRF formulation incorporates intensity information from the original images, the model parameters have to be re-adjusted whenever the image dataset is changed. In contrast, the proposed Post-DAE is agnostic to image intensity, and only needs to be trained once.} %Note that we do not compare Post-DAE with other methods like \cite{Oktay2017,Ravishankar2017} which incorporate anatomical priors while training the segmentation method itself, since these are not post-processing strategies.

\textcolor{black}{\subsection{Training Post-DAE} Post-DAE is independent of the segmentation methods and it was trained separately from them. The model was implemented in Keras and trained for 150 epochs (the architecture and training details are included in the Supplementary Material). During training, we used the mask degradation strategy described in Section \ref{sec:maskdegradation}. At test time, we took the segmentation masks generated by the baseline segmentation methods, and post-processed them by simply passing them through the DAE.}

\begin{table*}[t!]
%% increase table row spacing, adjust to taste
\renewcommand{\arraystretch}{1.4}
\caption{Mean and standard deviation for post-processing random forest predictions. The numbers in bold indicates that Post-DAE outperforms the other methods (no post-processing and CRF) with statistical significance \textcolor{black}{according to Wilcoxon test with Bonferroni correction.}}
%(p-value $<$ 0.05 according to Wilcoxon test). }
\label{RF_results}
\centering

\begin{tabular}{|c|c|c|c|c|c|c|c|c|}
\hline
\multicolumn{2}{|c|}{} &  \multicolumn{6}{c|}{JRST Chest X-ray}  & Sunnybrook Cardiac MR\\
\cline{3-9}
\multicolumn{2}{|c|}{Segmentations} &  \multicolumn{5}{c|}{Multi-class} & Binary & Binary\\
\cline{3-9}
\multicolumn{2}{|c|}{} &  Depth: 8 & Depth: 12 & Depth: 16 & Depth: 20 & Full model &Full model & Full model \\
\hline
\multirow{6}{*}{\textbf{Dice}}
&RF & 0.858 & 0.913 & 0.936 & 0.949 & 0.956  & 0.781&0.46\\
&&  (0.042) &  (0.033)&(0.029)&(0.029)&(0.028)&(0.070)&(0.24)\\
\cline{2-9}
& RF + CRF & 0.860 & 0.914 & 0.937 & 0.950 & 0.956  & 0.795&0.44\\
&& (0.042) & (0.032) & (0.029) & (0.029) &(0.027)&(0.074)&0.25\\
\cline{2-9}
& RF + DAE &  \textbf{0.922} &  \textbf{0.943} &  \textbf{0.948} &  0.951 &  0.951 & \textbf{0.865}&\textbf{0.47}\\
&&  \textbf{(0.024)} &  \textbf{(0.020)}& \textbf{(0.018)}& (0.018)&(0.019)& \textbf{(0.056)}&\textbf{(0.25)}\\

\hline
& RF& 102.26& 96.00 & 88.70 & 77.45&72.28 & 91.41&27.73\\
&&  (11.68)&(14.31) &(14.04)&(12.98)&(14.07)&(17.52)&(9.89)\\
\cline{2-9}
\textbf{Hausdorff Distance}

& RF+CRF & 101.17 &92.97 & 81.26 & 74.51 &67.29  &80.45&26.87\\
\textbf{(HD)} 
&& (12.94) &(14.72)&(12.73)&(13.10)&(13.53))&(22.28)&(10.03)\\
\cline{2-9}
& RF +DAE & \textbf{63.73} &  \textbf{60.72} & \textbf{62.47} & \textbf{62.95} &  \textbf{60.69} & \textbf{32.01}&\textbf{23.60}\\
&&   \textbf{(11.85)}&  \textbf{(12.20)}& \textbf{(15.16)}& \textbf{(16.84)}& \textbf{(14.12)}&  \textbf{(18.44)}&\textbf{(9.88)}\\
\hline
\end{tabular}
\end{table*}

\subsection{Baseline segmentation methods.} {\color{black} We trained binary and multi-class versions of two different segmentation models which produce masks of various qualities. \textcolor{black}{For the X-ray images, we tackled binary (lungs vs background) and multi-class (lungs, heart, background) segmentation. For the CMR images, we focus on binary LV segmentation. The} RF classifier was trained using intensity and texture features. For the binary segmentation, \textcolor{black}{we adopted a public implementation available online with default parameters\footnote{The source code and a complete description of the method is publicly available online at: \url{https://github.com/dgriffiths3/ml_segmentation}} which produces acceptable segmentation masks. }\textcolor{black}{It uses Haralick \cite{haralick1973textural} features which are based on gray level co-ocurrency in image patches}. For the multi-label segmentation, we apply a different implementation\footnote{Publicly available at: \url{https://github.com/biomedia-mira/oak2}} which has a better performance for multi-label predictions and has been used in \cite{glocker2013vertebrae}. This RF variant leverages randomized offset boxes for calculating average intensity and intensity difference features efficiently via integral images. For the multi-label segmentation using RF we produced segmentations of various qualities by using different tree depths at test time. The second method is a CNN based on UNet architecture \cite{RonnebergerUnet15} (see the \textcolor{black}{Supplementary Material} for a detailed description of the architecture and the training parameters such as optimizer, learning rate, etc.). \textcolor{black}{The UNet was implemented in Keras and trained in GPU using Dice loss function. To compare the effect of Post-DAE in different segmentation qualities, we save the UNet model every 5 epochs during training, and predict segmentation masks for the test fold using all these models.} We compared post-processing with Post-DAE and CRF, reporting results for all these cases.}

\section{Results and discussion}
Figure \ref{fig:qualitativeResults} shows some visual examples while Table \ref{RF_results} and Figure \ref{fig:quantitativeResults} summarize the quantitative results obtained when post-processing segmentations produced by a RF classifier and a UNet. \textcolor{black}{ Our best results are in line with those obtained for other deep learning based state-of-the-art methods. For JSRT, recent works \cite{frid2018improving,dai2018scan,novikov2018fully,mansilla2020} report average Dice values for lung and heart ranging from 0.943 \cite{mansilla2020} to 0.965 \cite{frid2018improving}. For the Sunnybrook dataset, recent works \cite{zheng20183,curiale2019automatic,chen2019fr,stough2018ventricular}} report average Dice ranging from 0.88 \cite{zheng20183} to 0.93 \cite{chen2019fr} for LV segmentation.
%{\color{black}(see the video in the Sup. Mat. for more visual results)}. 
Both figures show the consistent improvement achieved when using Post-DAE as a post-processing step, specially in \textcolor{black}{low quality segmentation masks like those obtained by the binary RF model, the multiclass RF considering incomplete tree depths and the UNet trained only for a few epochs}. In these cases, substantial improvements are obtained in terms of Dice coefficient and Hausdorff distance (HD), by bringing the erroneous segmentation masks into an anatomically feasible space. \textcolor{black}{In case of segmentations that are already of good quality (like {\color{black}multi-class RF} or the UNet trained until convergence), Post-DAE significantly improves the HD, by erasing spurious segmentations  that remain even in well trained models, like holes in the lung or small isolated blobs. When compared with CRF post-processing, Post-DAE significantly outperforms the baseline in the context of anatomical segmentation.} \textcolor{black}{In terms of running time, the CRF model takes} 1.3 seconds \textcolor{black}{while Post-DAE takes 0.76 seconds in a Intel i7-7700 CPU.}
%, while Post-DAE takes 0.7 seconds in a Titan Xp GPU.

Scatter plots in Figure \ref{fig:scatter} {\color{black} show the change in terms of Dice (top) and HD (bottom) between initial segmentations before post-processing (x-axis) and after post-processing (y-axis), when comparing them with the ground-truth masks. 
In the Dice plots we observe how the green points tend to concentrate in the upper part of the diagonal, while orange crosses stick to it, indicating that Post-DAE improves the segmentations more than CRF. HD scatter plots should be read in the opposite way, i.e. the lower the better. Green points, corresponding to Post-DAE, concentrate in the bottom part while the orange crosses (CRF) tend to be over them, indicating that Post-DAE outperforms CRF also in terms of HD.
%For the Dice plots (upper row), we observe how the green points tend to concentrate in the upper part of the diagonal, while orange crosses stick to it, indicating that Post-DAE improves the segmentations more than CRF. 
%For the Sunnybrook dataset, Post-DAE tends to improve the segmentations while CRF does not seem to help. 
%The Hausdorff distance (HD) scatter plots should be read in the opposite way (i.e. the lower the better). Green values (corresponding to Post-DAE) concentrate in the bottom part while the orange crosses (CRF) tend to be over them.

We include additional experiments in the Supplementary Material showing that Post-DAE can be trained with unpaired segmentation masks. These segmentation masks could be annotated on different image modalities, come from a different dataset with the same image modality or even from segmentation-only datasets. This experiment highlights the fact that our method does not require image intensity information for training/testing, making it robust to domain shift.}
 %{\color{black} First column shows the results for the JRST binary masks, second column for JRST multi-class and the third column shows the results for Sunnybrook segmentations}
%Top row shows the results for multi-class segmentations and bottom row shows the results for binary masks. In both cases we can observe that Post-DAE improves the results, outperforming post-processing with a CRF.

\begin{figure*}[h!]
\begin{center}
   \includegraphics[width=0.9\textwidth]{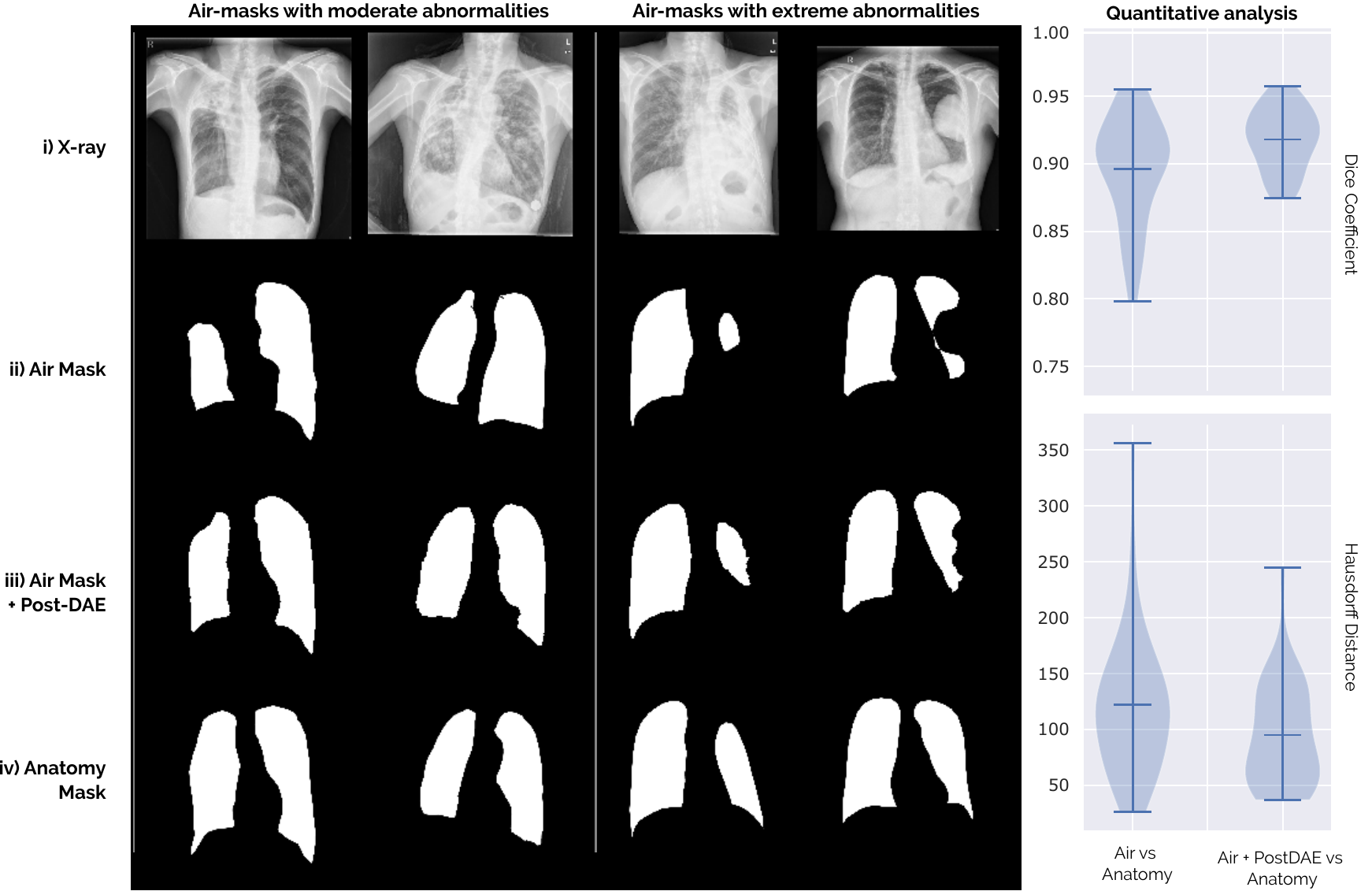}
    \caption{\textcolor{black}{Analysis for out-of-distribution segmentation masks presenting moderate and extreme abnormalities for tuberculosis patients from the Shenzen database. See Section \ref{sec:out-of-distribution}.A for a complete discussion about this experiment.}}
\label{fig:out-of-distribution}
\end{center}
\end{figure*}

\textcolor{black}{\subsection{Out-of-distribution segmentation masks and limitations}}
\label{sec:out-of-distribution}
\textcolor{black}{In this section we analyze the behaviour and limitations of Post-DAE when post-processing masks which are out-of-distribution. In this context, out-of-distribution cases could appear mainly due to two reasons: erroneous segmentations or pathological images.}

\textcolor{black}{ In the first case, erroneous masks may be generated by a segmentation method with low performance. See for example the masks in Figure \ref{fig:qualitativeResults}, obtained with the RF model or the UNet trained for only 5 or 10 epochs. These cases are represented in the scatter plots depicted in Figure \ref{fig:scatter} by the points with low Dice or high Hausdorff before post-processing. Post-DAE clearly improves erroneous segmentations in this scenario, increasing the Dice after post processing and/or reducing its Hausdorff distance. This improvement is explained by the way Post-DAE was trained: we degraded the ground truth masks by introducing similar errors (see Section \ref{sec:maskdegradation}) and force the DAE to reconstruct anatomically plausible segmentation masks. Since the test images are anatomically plausible as well, mapping erroneous segmentations to realistic ones improves the results.}
%In this case, Post-DAE provides a simple solution which works in practise.}

\textcolor{black}{The second scenario is related to abnormal cases. Big occlusions or deformed organs, possibly due to manifestations of a particular disease or radiological occlusions, could make these masks look entirely different from the anatomically plausible cases. To analyze the behaviour of our model in this scenario, we employed a different chest X-ray dataset containing patients diagnosed with tuberculosis. This dataset is a subset of the original Shenzhen database \cite{candemir2013lung,jaeger2013automatic} formed by 38 X-ray images exhibiting tuberculosis manifestations. Every image was annotated by two expert radiologists following different approaches to delineate the lungs as discussed in \cite{karargyris2016combination}. The first approach was to segment only the air cavity part of the lung field, i.e. segmenting only the dark part and ignoring lighter areas covered with fluid. We call these the \textit{air masks}. In the second approach, the annotator delineated the expected anatomy of the lungs, including occluded areas following a comparative approach by “mirroring” the normal lung field onto the abnormal one \cite{karargyris2016combination}. We call these \textit{anatomy masks}. Figure \ref{fig:out-of-distribution} shows examples for both types of segmentation masks (rows (ii) and (iv)). Note that those corresponding to the air approach might present moderate or extreme abnormalities (e.g. missing complete parts of the lung). We applied the Post-DAE model to the \textit{air masks} and analyzed its effect.}

\textcolor{black}{We used a Post-DAE model trained with the JSRT dataset, where the lung masks are mostly anatomically plausible since there are no big abnormalities or occlusions in the images. As expected, our method tends to map the air to the anatomy masks. However, note that when abnormalities are too extreme (see columns 3 and 4 in Figure \ref{fig:out-of-distribution}) the real anatomy can not be completely reconstructed. We quantified this experiment by measuring the Dice coefficient between the air and anatomy masks before and after post-processing the air masks with Post-DAE. The violin plots included in Figure \ref{fig:out-of-distribution} show that the post-processed air masks are significantly closer to the anatomy masks than the original ones, both in terms of Dice and Hausdorff metrics. This constitutes, at the same time, an advantage and a limitation of our approach: Post-DAE will transform the segmentation masks so that they look closer to the anatomically plausible ones used at training. These are important facts that must be considered when designing segmentation workflows which include Post-DAE. The same holds for problems different from anatomical segmentation. In scenarios like brain lesion or tumor segmentation, where shape and topology is not regular, the applicability of Post-DAE may be limited.}

\section{Conclusions}
In this work we have shown that denoising autoencoders can be used to render erroneous segmentations of different organs into anatomically plausible masks. Our method works as an independent post-processing step, allowing to incorporate anatomical priors into arbitrary segmentation methods. The provided experimental evaluation in the context of \textcolor{black}{binary and multi-class anatomical segmentation of X-ray and CMR images indicates that our method can deal with a variety of anatomical structures in different image modalities}. Moreover, Post-DAE does not use intensity information. Therefore, it can be trained with unpaired segmentation masks annotated on different image modalities or coming from segmentation-only datasets, making the method robust to domain shift.
\textcolor{black}{Post-DAE can be easily implemented, is fast at inference, can cope with arbitrary shape priors and is independent of the image modality and segmentation method.} In the future, we plan to explore the use of Post-DAE in the context of lesion segmentation \cite{roulet2019joint}, where the regions of interest are not as regular as anatomical structures.

%\textcolor{black}{One of the limitations of Post-DAE is related to data regularity. In case of anatomical structures like lung, heart or liver, even if we found high inter-subject variability, the segmentation masks are somehow uniform in terms of shape and topology. 
%Even pathological organs tend to have similar structure, which can be well-encoded by the DAE (specially if pathological cases are seen during training). 
%However, }

% or


\section*{Acknowledgments}
 \textcolor{black}{We thank Alexandros Karargyris, Sema Candemir and Stefan Jaeger for sharing the segmentation masks used in the out-of-distribution experiments.}

\ifCLASSOPTIONcaptionsoff
  \newpage
\fi

%\newpage

% \begin{figure*}[t!]
%   \includegraphics[width=\textwidth]{images/scatter.png}
%     \caption{OPTION 2 (WITHOUT CRF RESULTS): Scatter plots comparing Dice (left) and HD (right) before and after apply Post-DAE for all samples in the previous study. Top row shows the results for multi-class segmentation (average of per-class Dice) and bottom row shows the results for binary masks.}
% \label{fig:scatter}   
% \end{figure*}

\bibliographystyle{IEEEtran} 
\bibliography{library}
\newpage

\section*{Supplementary material}

\subsection{UNet details} The UNet model (see Table \ref{tab:unetarc}) receives a 1024x1024 gray image as input and was trained using the soft Dice loss \cite{milletari2016v}, batch size of 4, Adam optimizer with learning rate 1e-5 and the other parameters as by Keras default. We used data augmentation including random rotations, shifts, zoom and shear. We also used dropout for regularization, including a dropout layer after layer $L_5$ with keep probability p=0.5. For the multi-class UNet we used categorical cross-entropy loss and changed the initial learning rate to to 1e-4.

\begin{table}[h]
\caption{\label{tab:unetarc} Detailed description of the UNet architecture used as baseline model segmentation·}
\begin{center}
\resizebox{\columnwidth}{!}{\begin{tabular}{llllllll}
\hline
    &                 & \textbf{Kernel}\hspace{0.1 in}                      & \textbf{Stride} \hspace{0.1 in}                      &  \multicolumn{2}{c}{\#\textbf{Kernels}\hspace{0.1 in}}  &  \multicolumn{2}{c}{\textbf{NonLin}}  \\ \hline
        &                 &            &               &  Binary & MC  &  Binary & MC  \\ \hline
   
L1  & Conv            & (f:3,3)                     & (s:1,1)                     & (N:16)   & (N:16)    & ReLu& ReLu    \\
    & Conv            & (f:3,3)                     & (s:1,1)                     & (N:16)    & (N:16)   & ReLu    & ReLu\\ 
    & Max Pooling     & (f:2,2)                     & (s:2,2)                     &           &         \\ \hline
L2  & Conv            & (f:3,3)                     & (s:1,1)                     & (N:32)& (N:32)    & ReLu& ReLu    \\
    & Conv            & (f:3,3)                     & (s:1,1)                     & (N:32) & (N:32)   & ReLu& ReLu    \\ 
    & Max Pooling     & (f:2,2)                     & (s:2,2)                     &           &         \\ \hline
L3  & Conv            & (f:3,3)                     & (s:1,1)                     & (N:64)& (N:64)    & ReLu& ReLu    \\
    & Conv            & (f:3,3)                     & (s:1,1)                     & (N:64) & (N:64)   & ReLu& ReLu    \\ 
    & Max Pooling     & (f:2,2)                     & (s:2,2)                     &           &         \\ \hline
L4  & Conv            & (f:3,3)                     & (s:1,1)                     & (N:128) & (N:128)   & ReLu    & ReLu\\
    & Conv            & (f:3,3)                     & (s:1,1)                     & (N:128)  & (N:128)  & ReLu   & ReLu \\ 
    & Max Pooling     & (f:2,2)                     & (s:2,2)                     &           &         \\ \hline
L5  & Conv            & (f:3,3)                     & (s:1,1)                     & (N:256) & (N:256)   & ReLu  & ReLu  \\ 
    & Conv            & (f:3,3)                     & (s:1,1)                     & (N:256)  & (N:256)  & ReLu   & ReLu\\ \hline
L6  & UpConv          & (f:3,3)&                      (s:1,1)                     & (N:128)& (N:128)   & ReLu  & ReLu  \\
    & Conv            & (f:3,3)                 & (s:1,1) &                         (N:128) & (N:128)  & ReLu &ReLu   \\ 
    & Conv            & (f:3,3)                     & (s:1,1)                     & (N:128) & (N:128)   & ReLu  & ReLu  \\\hline
L7  & UpConv          & (f:3,3)                     & (s:1,1)                     & (N:64)  & (N:64)    & ReLu & ReLu   \\ 
    & Conv            & (f:3,3)                     & (s:1,1)                     & (N:64) & (N:64)     & ReLu  & ReLu  \\
    & Conv            & (f:3,3)                     & (s:1,1)                     & (N:64)     & (N:64)  & ReLu & ReLu   \\ \hline
L8  & UpConv          & (f:3,3)                     & (s:1,1)                     & (N:32)   & (N:32)    & ReLu & ReLu   \\
    & Conv            & (f:3,3)                     & (s:1,1)                     & (N:32)    & (N:32)   & ReLu & ReLu   \\ 
    & Conv            & (f:3,3)                     & (s:1,1)                     & (N:32)   & (N:32)    & ReLu & ReLu   \\\hline
L9  & UpConv          & (f:3,3)                     & (s:1,1)                     & (N:16) & (N:16)    & ReLu & ReLu   \\ 
    & Conv            & (f:3,3)                     & (s:1,1)                     & (N:16)   & (N:16)  & ReLu & ReLu   \\
    & Conv            & (f:3,3)                     & (s:1,1)                     & (N:16)  & (N:16)   & ReLu & ReLu   \\ \hline
  L10  & Conv            & (f:3,3)                     & (s:1,1)                     & (N:2)& (N:3)     & ReLu& ReLu    \\
 & Conv            & (f:1,1)                     & (s:1,1)                     & (N:1) & (N:3)    & Sigmoid &SoftMax\\\hline

\end{tabular}}
\end{center}
\end{table}

\subsection{Post-DAE} Post-DAE (see Table \ref{tab:postdae_arc}) receives a 1024x1024 segmentation as input. The network was also trained to minimize the Dice loss function using Adam Optimizer. We used learning rate of 0.0001, batch size of 15 and 150 epochs. The multi-class Post-DAE implementation receives a one-hot encoded segmentation of size 1024x1024x3 segmentation as input. Because of memory restrictions, in this case we reduced the batch size to 8.

\begin{table}[t]
\caption{\label{tab:postdae_arc} Detailed architecture of the simple denoising auto encoder model used to implement the proposed Post-DAE.}
\begin{center}
\resizebox{\columnwidth}{!}{\begin{tabular}{llllllll}
\hline
    &        & \textbf{Kernel}                & \textbf{Stride}         & \multicolumn{2}{c}{\#\textbf{Kernels}} & \multicolumn{2}{c}{\textbf{NonLin}}  \\ \hline
        &        &                &                 & Binary & MC & Binary &MC \\ \hline
    $L_1$  & Conv   & (f:3,3)               & (s:2,2)               & (N:16)&  (N:16)  & ReLu  & ReLu  \\
    & Conv   & (f:3,3)               & (s:1,1)               & (N:16)&(N:16    & ReLu &  ReLu   \\ \hline
L2  & Conv   & (f:3,3)               & (s:2,2)               & (N:32) & (N:32)  & ReLu & ReLu    \\
    & Conv   & (f:3,3)               & (s:1,1)               & (N:32) & (N:32)  & ReLu &  ReLu  \\ \hline
L3  & Conv   & (f:3,3)               & (s:2,2)               & (N:32) & (N:32)  & ReLu &   ReLu\\
    & Conv   & (f:3,3)               & (s:1,1)               & (N:32) &(N:32)   & ReLu  & ReLu  \\ \hline
L4  & Conv   & (f:3,3)               & (s:2,2)               & (N:32) & (N:32)  & ReLu  &   ReLu\\
    & Conv   & (f:3,3)               & (s:1,1)               & (N:32) & (N:32)  & ReLu  &   ReLu\\ \hline
L5  & Conv   & (f:3,3)               & (s:2,2)               & (N:32) & (N:32)  & ReLu  &  ReLu \\ \hline
L6  & FC     & \multicolumn{1}{c}{-} & \multicolumn{1}{c}{-} & (N:512) & (N:1024) & None  &None  \\ \hline
L6  & FC     & \multicolumn{1}{c}{-} & \multicolumn{1}{c}{-} & (N:1024) &(N:4096) & Relu   &Relu \\ \hline
L8  & UpConv & (f:3,3)               & (s:1,1)               & (N:16) & (N:16)  & ReLu  &   ReLu\\
    & Conv   & (f:3,3)               & (s:1,1)               & (N:16)  & (N:16) & ReLu  &   ReLu\\ \hline
L9  & UpConv & (f:3,3)               & (s:1,1)               & (N:16) & (N:16)  & ReLu  &  ReLu \\
    & Conv   & (f:3,3)               & (s:1,1)               & (N:16) & (N:16)  & ReLu  &   ReLu\\ \hline
L10 & UpConv & (f:3,3)               & (s:1,1)               & (N:16) & (N:16)  & ReLu  &  ReLu \\
    & Conv   & (f:3,3)               & (s:1,1)               & (N:16) & (N:16)  & ReLu  &   ReLu\\ \hline
L11 & UpConv & (f:3,3)               & (s:1,1)               & (N:16) & (N:16)  & ReLu  &   ReLu\\
    & Conv   & (f:3,3)               & (s:1,1)               & (N:16) & (N:16)  & ReLu  &  ReLu \\ \hline
L12 & UpConv & (f:3,3)               & (s:1,1)               & (N:16)& (N:16)   & ReLu  &  ReLu \\
    & Conv   & (f:3,3)               & (s:1,1)               & (N:1) & (N:3)   & Sigmoid & SoftMax \\ \hline

\end{tabular}}
\end{center}
\end{table}

\subsection{Additional experiments for segmentation-only datasets}
\begin{figure*}[t!]
\begin{center}
   \includegraphics[width=0.9\textwidth]{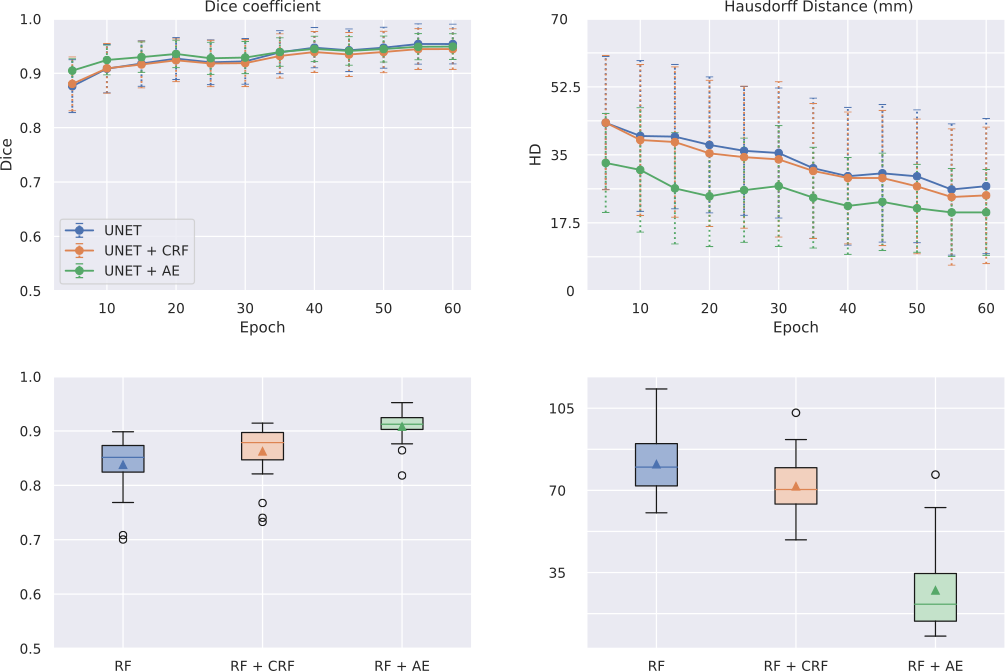}
    \caption{Quantitative evaluation of the proposed method in a new data-set, which was not seen by the Post-DAE. {\color{black} Top row shows mean and standard deviation for post-processing UNet predictions on the test fold at different training stages (from 5 epochs to convergence). We use Dice coefficient and Hausdorff distance to measure the segmentation quality. Bottom row show results for post-processing the random forest predictions. The triangle in the box indicates the mean value.} }
\label{fig:Montgomery}   
\end{center}
\end{figure*}

 We performed an extra experiment aiming to show that it is possible to use annotations from a different dataset to train Post-DAE. 
 %These labels correspond to unpaired segmentation masks that could be annotated on different image modalities or come from segmentation-only datasets. This is specially important: the fact that our method does not require image intensity information for training/testing makes it robust to domain shift.
We used the DAE trained with JSRT database to post-process results obtained with the binary UNet for the Montgomery County X-ray Set \cite{jaeger2014two}, a different chest X-ray dataset with manual lung annotations. X-ray images in this dataset were acquired from the tuberculosis control program of the Department of Health and Human Services of Montgomery County, MD, USA. This set contains 138 posterior-anterior x-rays, which were divided in 100 for training, 14 for validation and 24 for testing. The size of the X-rays is either 4020x4892 or 4892x4020 pixels. All images were downsampled to 1024x1024 in our experiments, padded with 0's to obtain a 1:1 aspect ratio and rigidly aligned. With this dataset, we trained two segmentation models: a Random Forest and a UNet architecture (saving its output every 5 epochs during training), predicting segmentation masks for the test fold using all these models. These masks were then post-processed using Post-DAE. 
Figure \ref{fig:Montgomery} shows the results for this experiment, where unpaired annotations coming from a different dataset are used to train Post-DAE. It can be observed how our method improves the segmentation quality even when the annotations used to train Post-DAE are coming from a different dataset.

%\bibliographystyle{IEEEtran}


%\vfill


\end{document}